\documentclass{article}

\usepackage{PRIMEarxiv}

\usepackage[utf8]{inputenc} 
\usepackage[T1]{fontenc}    
\usepackage{hyperref}       
\usepackage{url}            
\usepackage{booktabs}       
\usepackage{amsfonts}       
\usepackage{nicefrac}       
\usepackage{microtype}      
\usepackage{lipsum}
\usepackage{fancyhdr}       
\usepackage{graphicx}       
\graphicspath{{media/}}     
\usepackage{amsmath}
\usepackage{amssymb}
\usepackage{mathtools}
\usepackage{bm}
\usepackage{float}
\usepackage{multirow}
\usepackage{array}
\usepackage{xcolor}

\title{Industrial Tokenization for LLM-Based Health Intelligence: A Federated Architecture for Industrial Evidence Integration
}

\author{
Deshui Li$^{*}$ \\
Mingyang Smart Energy Co., Ltd. \\
Zhongshan, China \\
\texttt{lideshui@mywind.com.cn}
\and
Xiao-Ming Yuan$^{*,\dagger}$ \\
Mingyang Smart Energy Co., Ltd. \\
Zhongshan, China \\
ORCID: \href{https://orcid.org/0009-0002-0977-585X}{0009-0002-0977-585X} \\
\texttt{yuanxiaoming@mywind.com.cn} \\
\texttt{eeyxm@outlook.com}
\and
Zishun Wang \\
Mingyang Smart Energy Co., Ltd. \\
Zhongshan, China \\
\texttt{wangzishun@mywind.com.cn}
}

\begin{document}
\maketitle

\begingroup
\renewcommand{\thefootnote}{\fnsymbol{footnote}}
\footnotetext[1]{Deshui Li and Xiao-Ming Yuan contributed equally to this work.}
\footnotetext[2]{Corresponding author: Xiao-Ming Yuan.}
\endgroup

\begin{abstract}
Industrial health management increasingly relies on heterogeneous information sources, including condition monitoring systems, supervisory control and data acquisition systems, maintenance records, inspection results, and prognostic models. Although large language models provide new opportunities for cross-source reasoning and decision support, industrial data and analytical outputs differ substantially in structure, temporal resolution, physical meaning, and reliability. Directly integrating such heterogeneous information into a monolithic model may therefore reduce interpretability, traceability, and adaptability to changes in equipment configuration and available data sources.
This paper introduces \emph{Industrial Tokenization}, a conceptual interface for transforming source-specific analytical outputs into structured and machine-interpretable units of industrial evidence, termed \emph{Industrial Tokens}. Unlike numerical tokens used to encode raw time-series data, Industrial Tokens represent domain-grounded evidence together with its source, temporal scope, operating context, analytical meaning, quality or confidence information, and provenance.
Based on this concept, a federated industrial architecture is proposed in which heterogeneous analytical subsystems retain autonomy in data processing and model selection while exposing standardized Industrial Tokens to a central reasoning layer. As an initial implementation, this study presents an end-to-end DiagnosisToken pathway based on vibration-diagnostic outputs, rule-based event aggregation, structured textual token generation, and LLM-based interpretation. Other Industrial Tokens, such as SCADA-based condition-monitoring tokens, maintenance tokens, and prognostic tokens, are
reserved as future extensions within the proposed architecture. The proposed framework positions \emph{Industrial Tokenization} as a stable semantic interface between domain-specific industrial intelligence and LLM- or agent-based reasoning, rather than as another method for encoding raw industrial data.
\end{abstract}

\keywords{
Industrial Tokenization \and
Industrial Artificial Intelligence \and
Fault Diagnosis \and
Large Language Models \and
Industrial Information Integration
}

\section{Introduction}
The increasing deployment of sensors, control systems, and digital platforms has resulted in the continuous accumulation of heterogeneous industrial data. Typical information sources include vibration signals collected by condition monitoring systems (CMS), operating and condition variables recorded by supervisory control and data acquisition (SCADA) systems, control states obtained from programmable logic controllers (PLCs), alarm records, maintenance logs, inspection results, and prognostic outputs. These sources describe different aspects of equipment operation, degradation, and maintenance history.

Industrial monitoring and diagnosis have traditionally been developed around individual data sources. Vibration signals directly reflect the dynamic behavior of rotating machinery and are widely used for bearing, gearbox, and generator fault diagnosis. SCADA data provide lower-frequency operating, thermal, and subsystem condition information. PLC records describe discrete control actions and component states, such as motor activation, valve switching, protection logic, and subsystem transitions. Because these sources differ in physical meaning, data structure, and analytical requirements, they are commonly processed using separate domain-specific models.

With the increasing availability of industrial data, research has gradually expanded from source-specific monitoring toward multisource and multimodal diagnosis. Vibration, operational, thermal, acoustic, image, maintenance, and textual information have been combined to support more comprehensive equipment health assessment. More recently, the development of large language models (LLMs) and multimodal foundation models has motivated new attempts to use general-purpose AI for industrial knowledge retrieval, maintenance assistance, report generation, diagnosis, and decision support.

However, real industrial data are substantially more heterogeneous than is typically represented in benchmark studies. Even within the same data source, measurement configurations and data definitions may vary across equipment types, product generations, and suppliers. In vibration monitoring, sampling frequency, sampling duration, sensor location, orientation, manufacturer, physical unit, and acquisition system may all differ. In SCADA systems, different product platforms may adopt different subsystem designs, variable definitions, control strategies, and data schemas. PLC configurations may also change across production batches because of differences in controllers, communication protocols, variable addresses, and component suppliers.

Such heterogeneity creates a scalability problem for industrial intelligence. A diagnostic or prognostic model may perform reliably within the equipment configuration for which it was developed, particularly when detailed physical knowledge is incorporated. Nevertheless, the same model may be difficult to reuse when the measurement configuration, subsystem design, equipment generation, or available data sources change. Continuously expanding a monolithic model to accommodate every new sensor, subsystem, and product variant can lead to repeated redevelopment and poor maintainability. Directly providing heterogeneous raw data to a general-purpose large model may also result in semantic ambiguity, loss of physical meaning, weak traceability, and unreliable reasoning.

These limitations raise a broader architectural question: how can source-specific models, multimodal models, and LLM-based reasoning be connected without forcing heterogeneous industrial data and analytical procedures into a single monolithic system?

To address this question, this paper introduces \emph{Industrial Tokenization} within a \emph{Federated Industrial Architecture}. Industrial Tokenization transforms the outputs of heterogeneous analytical modules into structured and machine-interpretable evidence units termed \emph{Industrial Tokens}. Rather
than representing raw signal segments or generic numerical embeddings, Industrial Tokens describe domain-grounded evidence together with its source, temporal scope, operating context, analytical meaning, quality or confidence information, and provenance.

The proposed federated architecture organizes analytical modules as autonomous but interoperable subsystems. In this context, ``federated'' does not refer to federated learning. Instead, each subsystem independently processes its own data, selects suitable analytical methods, and produces specialized evidence.
The resulting evidence is standardized through Industrial Tokenization and provided to a central LLM- or agent-based reasoning layer. This design preserves the independence of specialized industrial models while providing a relatively stable semantic interface for cross-source interpretation.

The architecture is particularly relevant to evolving industrial products. When sensors, subsystem designs, equipment variants, or analytical models change, the corresponding source-specific module and token interface can be adapted without requiring the entire reasoning framework to be redesigned. The framework therefore does not eliminate product-specific model development; rather, it allows heterogeneous industrial models to evolve independently while remaining interoperable at the evidence level.

The main contributions of this paper are summarized as follows:

\begin{itemize}
    \item Industrial Tokenization is introduced as a conceptual interface for transforming source-specific analytical outputs into structured, domain-grounded, and traceable industrial evidence.

    \item A Federated Industrial Architecture is proposed in which heterogeneous analytical subsystems retain autonomy while exposing standardized Industrial Tokens to a central reasoning layer.

    \item An initial end-to-end DiagnosisToken pathway is implemented using vibration-based diagnostic outputs, rule-based event aggregation, structured textual token generation, and LLM-based interpretation.
\end{itemize}

The remainder of this paper is organized as follows. Section~\ref{sec:related_work} reviews related concepts and identifies the missing interface for industrial evidence. Section~\ref{sec:induatdy_tokenization} introduces Industrial Tokenization and distinguishes Industrial Tokens from raw-data tokens and conventional analytical outputs. Section~\ref{sec:federated_industrial_architecture} presents the Federated Industrial Architecture. Section ~\ref{sec:end_to_end_diagnosis_token} presents an initial implementation of the DiagnosisToken pathway. Section~\ref{sec:future_work} discusses the implications, limitations, and future development of the framework, followed by the conclusion in Section~\ref{sec:conclusion}.

\section{Related Concepts and Research Gap}
\label{sec:related_work}

\subsection{Source-Specific and Multimodal Industrial Monitoring}

Industrial condition monitoring has traditionally been developed around
source-specific analytical methods. Vibration signals are widely used for the
condition assessment and fault diagnosis of rotating machinery because they
directly reflect the dynamic response of mechanical components. Representative
applications include vibration-based diagnosis of pumps, electrical motors,
and planetary gearboxes \cite{WANG2006176,1546063,FENG20124919}. These methods
generally rely on signal-processing, physical modeling, or data-driven
classification to identify fault-related patterns from a specific monitoring
source.

SCADA data provide another important source of operational information,
particularly for large industrial assets such as wind turbines. SCADA-based
condition-monitoring methods commonly use variables collected at relatively
low sampling rates to model normal operating behavior and detect deviations.
For example, spatio-temporal features extracted from wind-turbine SCADA data
have been combined using convolutional and recurrent neural networks for
condition assessment \cite{KONG2020760}. Although such methods may fuse
multiple variables or temporal features, they still operate mainly within a
single monitoring source.

To obtain a more comprehensive description of equipment condition, recent
studies have explored multi-sensor and multimodal fusion. Condition-based
maintenance research has long recognized the value of combining information
from multiple sensors and analytical processes \cite{JARDINE20061483}. More
recent work has integrated vibration signals with thermal images for gearbox
fault diagnosis \cite{ZHANG2024108236}, while multimodal feature fusion has
also been investigated for bearing diagnosis under varying operating
conditions \cite{10960316}. These studies demonstrate that complementary
modalities can improve fault representation and diagnostic robustness.

However, existing source-specific and multimodal methods mainly focus on
combining raw data, extracted features, or model predictions within a
particular diagnostic task. In practical industrial systems, different
analytical subsystems may still produce results with different formats,
terminologies, temporal resolutions, and levels of detail. Therefore, even
when multiple monitoring modules describe the same equipment, their analytical
outputs may not be directly suitable for unified interpretation and
system-level reasoning.

\subsection{Time-Series Tokenization and LLM-Based Industrial Intelligence}

The development of Transformer-based models has motivated increasing research
on representing time-series data as token sequences. A time-series tokenizer
can divide or transform continuous numerical signals into model-compatible
representations before they are processed by a Transformer. For rotating
machinery fault diagnosis, a time-series tokenizer has been combined with a
Time Series Transformer to generate token sequences from one-dimensional
signals \cite{JIN2022379}. Feature-tokenizer structures have also been
introduced for fault diagnosis in marine propulsion systems
\cite{11580333}. More recently, heterogeneous signal embedding has been
investigated as a plug-and-play module for unified fault diagnosis foundation
models, where signals with different types, sampling rates, and lengths are
projected into a unified representation space \cite{LI2025103277}.

More broadly, recent reviews have summarized the rapid development of
time-series large language models, tokenization strategies, and deep
time-series foundation models \cite{10856008,11509648}. These studies show that
tokenization is becoming an important mechanism for connecting numerical
sequences with Transformer-based or language-model-inspired architectures.
Their primary objective is generally to enable a model to learn patterns
directly from raw or lightly processed time-series data.

In parallel, LLMs have begun to be incorporated into industrial monitoring, fault diagnosis, and maintenance workflows. Domain-specific knowledge bases have been used to support LLM-based operation and maintenance assistance \cite{10336112}. Large visual and language models have also been combined for industrial visual monitoring and maintenance \cite{10557154}. In addition, LLM-TSFD introduces an LLM-based human-in-the-loop framework for industrial time-series fault diagnosis, in which the LLM participates in data-pipeline management, model management, causal correction, and diagnostic decision-making \cite{ZHANG2025125861}.

Furthermore, MaintAGT explores a multimodal large model for intelligent maintenance by introducing a signal-to-text module that converts monitoring signals into textual descriptions and integrating them with domain-specific maintenance knowledge for LLM-based reasoning \cite{HE202412481}. Such studies demonstrate the potential of transforming industrial signals
into language-compatible representations to enable large-model-based diagnosis and maintenance assistance.

These studies establish important foundations for time-series tokenization, heterogeneous signal representation, and LLM-based industrial intelligence. Nevertheless, most existing approaches focus on raw-data representation, feature tokenization, signal embedding, model adaptation, knowledge retrieval, or LLM-assisted decision-making. They do not necessarily address how the outputs of heterogeneous and independently developed industrial analytical subsystems can be converted into a relatively unified form for central interpretation and fusion.

\subsection{The Missing Industrial Evidence Interface}

The studies reviewed above address two important aspects of industrial intelligence. Source-specific and multimodal monitoring methods extract diagnostic information from heterogeneous industrial data, while time-series tokenization and LLM-based methods provide new ways to process numerical data, domain knowledge, and diagnostic tasks \cite{JARDINE20061483,ZHANG2024108236,JIN2022379,ZHANG2025125861}.

However, comparatively limited attention has been paid to the interface between autonomous industrial analytical subsystems and a central LLM. In a practical monitoring system, vibration-diagnosis models, SCADA-based condition-monitoring models, maintenance-analysis modules, and prognostic models may be developed independently. These subsystems can differ in their input data, internal mechanisms, output structures, update frequencies, and domain-specific terminology.

Directly integrating their raw data or internal model representations may be impractical because it would require the central model to understand every source-specific data structure and analytical process. Likewise, directly combining their original outputs may be difficult when the results are
expressed using incompatible labels, numerical indicators, reports, or decision rules.

This motivates the need for an intermediate industrial evidence interface that can preserve the main conclusions, supporting indicators, temporal information, and confidence statements produced by each subsystem, while expressing them in a form that can be interpreted by an LLM. To address this gap, this study introduces Industrial Tokenization, which transforms source-specific analytical outputs into relatively unified, language-based Industrial Tokens for subsequent interpretation and fusion.

\section{Industrial Tokenization}
\label{sec:induatdy_tokenization}
\subsection{Definition of an Industrial Token}
Industrial monitoring data are naturally heterogeneous. This heterogeneity arises not only from different data sources, but also from differences in industries, equipment types, machine structures, product generations, and suppliers. Even similar machines, such as generators, motors, or compressors, may use different sensors, acquisition systems, variable definitions, and data formats. Therefore, their monitoring data and analytical results cannot be assumed to share a unified form. To address this issue, we aim to transform analytical results from different industries and machines into a unified, language-based intermediate representation.

Although English, Mandarin, French, and German differ in vocabulary and grammatical structure, the meanings expressed in these languages can still be understood at a common semantic level. Language can describe an industrial object using both qualitative and quantitative information. A large language model (LLM) can therefore interpret the condition and severity of a target machine from such descriptions. When long-term information is included, the LLM can further aggregate and fuse heterogeneous, source-specific analytical results, form an overall assessment of equipment health, and generate a corresponding description of the machine condition.

Inspired by this idea, we propose \emph{Industrial Tokenization}, which transforms the analytical results produced by different industrial equipment and models into a relatively unified language representation. These representations, termed \emph{Industrial Tokens}, can then be provided to an LLM or agent for cross-source evidence fusion, equipment health assessment, and condition description.

An Industrial Token is therefore defined as a relatively standardized, language-based representation of a source-specific industrial analytical result, which can be interpreted and fused by a central LLM or agent.

\subsection{Rule-Based Industrial Tokenization}

At the current stage, source-specific tokenization is implemented using expert-defined rules. These rules map the outputs of diagnostic and condition-monitoring models into relatively unified language descriptions. The resulting rule-based Industrial Tokens provide the initial representation used for cross-source evidence fusion.

As more industrial cases, expert revisions, and validated token descriptions are accumulated, more automated tokenization methods may be investigated in future work.
\subsection{Distinction from Raw-Data Tokens and Analytical Outputs}

Raw-data tokenization and Industrial Tokenization can be viewed as two complementary attempts to connect non-linguistic information with AI models. Raw-data tokenization is commonly developed from the perspective of AI for Science, where numerical sequences, physical measurements, or scientific data are transformed into tokens that can be processed by foundation models. Its main purpose is to enable an AI model to directly learn representations and patterns from raw or lightly processed data.

Industrial Tokenization approaches the same interface problem from the industrial side. Its target is not limited to a specific dataset, machine, or application case. Instead, it aims to accommodate information from different industries, equipment types, suppliers, monitoring systems, and analytical methods. Because industrial systems are naturally heterogeneous and may also contain overlapping or redundant information, Industrial Tokenization focuses on transforming source-specific analytical results into relatively unified, sufficiently descriptive, and LLM-readable representations.

Raw-data tokens and Industrial Tokens may therefore coexist within the same architecture. Raw-data tokens can be used inside an individual subsystem to support signal analysis, pattern recognition, or state estimation. The output of that subsystem can then be transformed into an Industrial Token and provided to the central LLM for cross-source fusion and overall equipment health assessment. In this sense, the output of the source-specific analytical module provides a handshake between raw-data tokenization and Industrial Tokenization.

A conventional analytical output usually describes the result of one model, one data source, or one component. By contrast, Industrial Tokenization requires the outputs associated with the same target equipment to follow a relatively unified and sufficiently informative form, so that the central LLM can understand, compare, aggregate, and fuse them. Industrial Tokens do not replace source-specific analytical outputs; they make these local results suitable for system-level health assessment.

In this paper, an Industrial Token refers to a structured textual representation used as an intermediate medium between a source-specific analytical subsystem and the central LLM. It does not refer to the internal subword tokens, token vocabulary, or tokenization mechanism of the LLM itself. The internal tokenization process of the LLM is outside the scope of this study.

\section{Federated Industrial Architecture}
\label{sec:federated_industrial_architecture}
\begin{figure}
    \centering
    \includegraphics[width=\linewidth]{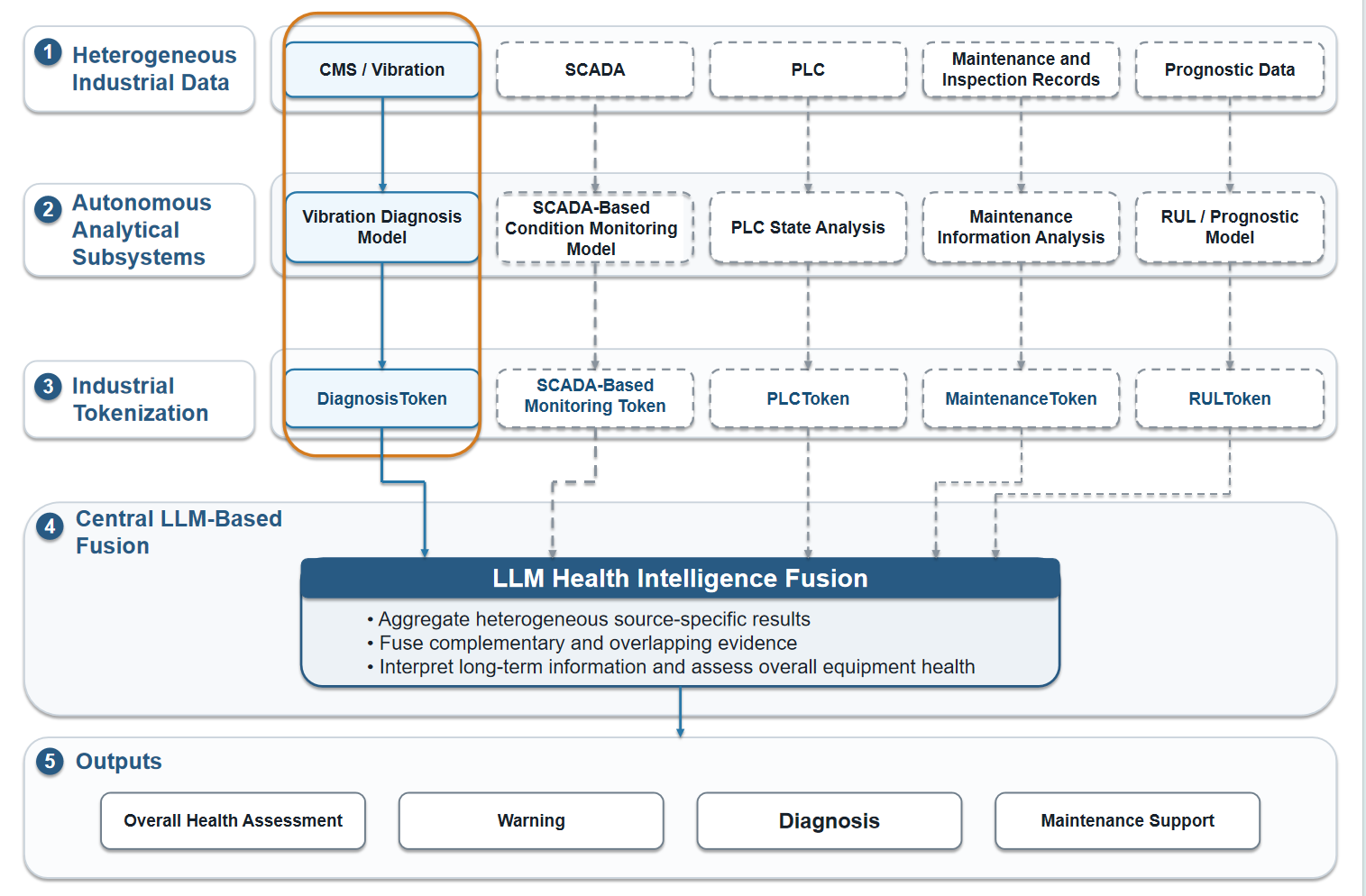}
    \caption{Federated Industrial Architecture for Industrial Tokenization. Heterogeneous industrial data are processed by autonomous analytical subsystems and transformed into Industrial Tokens before being provided to a central LLM-based fusion layer. The solid path denotes the implemented end-to-end DiagnosisToken pathway, whereas the dashed paths represent future analytical subsystems and Industrial Token extensions.}
    \label{fig:federated_architecture}
\end{figure}

The proposed Federated Industrial Architecture consists of autonomous analytical subsystems, a unified Industrial Token interface, and a central LLM-based fusion layer. As illustrated in Fig.~\ref{fig:federated_architecture}, heterogeneous industrial data are first processed by source-specific models. Their analytical outputs are then transformed into Industrial Tokens and provided to the central LLM for cross-source fusion and overall equipment health assessment.

The proposed architecture provides a general framework for integrating multiple autonomous industrial analytical subsystems. In the present study, the framework is instantiated through an end-to-end DiagnosisToken pathway.

\subsection{Autonomous Analytical Subsystems}
In the proposed architecture, each industrial data source is processed by an independent analytical subsystem. For example, vibration data may be analyzed by a fault-diagnosis model, SCADA data may be processed by a condition-monitoring model, and maintenance records may be interpreted by a maintenance-information analysis module. These subsystems may use different data structures, physical laws, analytical methods, and update frequencies.

The federated architecture does not require the internal models to be homogeneous. A subsystem may use signal-processing methods, physics-based models, expert rules, machine-learning models, or their combinations. Each subsystem is responsible for producing a local analytical result for its own data source or equipment component. Therefore, a model can be independently added, updated, replaced, or removed without redesigning the other analytical subsystems.
\subsection{Unified Token Interface}
Although the analytical subsystems remain independent, their outputs need a relatively unified form before they can be interpreted and fused by the central LLM. Industrial Tokenization provides this intermediate interface by transforming source-specific analytical results into language-based Industrial Tokens.

The unified interface does not require all Industrial Tokens to contain identical information. Instead, it requires them to provide sufficiently clear and understandable descriptions of the local analytical results. Such descriptions may include qualitative conclusions, quantitative indicators, temporal information, and other information required to explain the condition of the corresponding subsystem or component.

At the current stage, this transformation can be implemented using expert-defined rules. Different analytical outputs are mapped into relatively consistent language descriptions, which form the initial Industrial Tokens. The interface can later be extended as more industrial cases and validated token descriptions are accumulated.

\subsection{Central LLM-Based Fusion}
The central LLM-based layer receives Industrial Tokens rather than directly processing all heterogeneous raw industrial data. Its main role is to aggregate and fuse the local results generated by the autonomous analytical subsystems. Because these results describe different aspects of the same target equipment, the central layer can combine complementary or overlapping evidence and form a more comprehensive understanding of equipment health.

When long-term Industrial Tokens are available, the central LLM can also consider historical changes in diagnostic and condition-monitoring results. Based on the fused evidence, it can generate an overall health assessment, warning information, diagnostic descriptions, and maintenance-support information. Therefore, the central LLM acts as the integration layer of the Federated Industrial Architecture, while the source-specific subsystems retain responsibility for professional data analysis.

In the current implementation, only the DiagnosisToken pathway is instantiated. Therefore, the LLM-based layer is demonstrated through single-source diagnostic evidence interpretation, while cross-source fusion is left for future work.

\section{Implementation of DiagnosisToken}
\label{sec:end_to_end_diagnosis_token}
This section presents an initial end-to-end implementation of DiagnosisToken. The implemented pathway starts from automated vibration-based fault diagnosis,
continues with rule-based aggregation of record-level diagnostic results, and ends with ChatGPT-based interpretation of the generated textual DiagnosisTokens. The purpose of this section is not to disclose the detailed fault-diagnosis algorithms, but to demonstrate how source-specific diagnostic outputs can be organized into structured textual representations and connected to an LLM.
\subsection{Automated Vibration-Based Fault Diagnosis}

An automated vibration-based diagnostic system was applied to the monitoring data collected from a wind farm. For each turbine, measurement configuration, and vibration record within the selected one-month period, the system automatically evaluated the signal condition and generated a record-level fault-diagnosis result.

Each result was associated with a timestamp, turbine identifier, component, measurement position, measurement direction, and sampling frequency. The record-level outputs were stored in a database and subsequently used as the input to the DiagnosisToken generation process.

The detailed vibration-diagnosis algorithms, the automated diagnostic framework, and its software implementation are outside the scope of this study. In the proposed architecture, the diagnostic system is treated as an autonomous, source-specific analytical subsystem, while the present work focuses on organizing and transforming its outputs into an LLM-readable intermediate representation.

\subsection{Rule-Based Event Aggregation and DiagnosisToken Generation}

The record-level diagnostic outputs were further processed using a rule-based event aggregation module. Instead of constructing one token for each vibration record, the module organized the results according to turbine, component, measurement position, measurement direction, sampling frequency, and analysis period. Each such combination represents an independent diagnostic object within the selected time window.

For each diagnostic object, the record-level results were first aggregated on a daily basis. The number of available records, distorted-signal records, and fault-indicating records was calculated for each day. A day was identified as a distortion day or fault day according to predefined count- and ratio-based rules.

The daily results were then summarized over the complete analysis period. The resulting event representation included data observability, the number of distortion and fault days, their frequencies over valid monitoring days, and the maximum numbers of consecutive distortion and fault days. These indicators describe not only whether an abnormal result occurred, but also its recurrence,
persistence, and temporal distribution.

A second set of rules converted the aggregated indicators into normal, warning, or critical states. Signal quality and fault diagnosis were evaluated separately. The signal-quality result was additionally used to qualify the reliability of the fault-diagnosis conclusion. For example, diagnostic conclusions generated during periods of severe signal distortion were marked as unreliable, whereas conclusions obtained from signals with good quality were considered valid.

The aggregated event, interpreted status, confidence statement, and maintenance-oriented description were finally converted into a structured textual representation. This representation constitutes the DiagnosisToken implemented in this study.

\subsection{LLM-Based Interpretation}

In this preliminary demonstration, the generated DiagnosisTokens were manually provided to ChatGPT through its web interface. ChatGPT did not directly access the raw vibration signals, the record-level diagnostic database, or the internal diagnostic algorithms. Instead, it interpreted the aggregated diagnostic evidence represented by the DiagnosisTokens.

Two representative cases were selected to illustrate this process. The first case represents persistent fault evidence obtained under normal signal quality,
whereas the second represents severe signal degradation that limits the reliability of the fault-diagnosis conclusion. The same prompt was used for both cases, requesting ChatGPT to summarize the equipment condition, identify the supporting evidence, explain the influence of signal quality on diagnostic confidence, and provide a maintenance-oriented interpretation.

\paragraph{Case 1: Persistent fault evidence under normal signal quality.}

The first anonymized DiagnosisToken described the main-bearing condition of turbine WT-A over a one-month analysis period. The data observability was 96.8\%, and the signal-quality status was classified as normal. Fault indications occurred on 22 valid monitoring days, corresponding to a fault frequency of 73.3\%, with a maximum consecutive duration of six days. The aggregated fault status was classified as critical, and the diagnostic conclusion was considered valid because no severe signal-quality problem was identified.

The textual DiagnosisToken provided to the LLM was formulated as follows:

\begin{quote}
Turbine WT-A, main bearing, left radial measurement position, was monitored over a one-month period with a data observability of 96.8\%. The signal-quality status was normal. Fault indications occurred on 22 valid monitoring days,
corresponding to a fault frequency of 73.3\%, with a maximum consecutive duration of six days. The fault status was classified as critical. Because the signal quality was acceptable, the diagnostic conclusion was considered valid. Priority monitoring and field inspection were recommended.
\end{quote}

The LLM was then asked to identify the monitored component, assess the persistence of the fault evidence, explain the confidence of the diagnostic conclusion, and provide an appropriate maintenance-oriented interpretation.


\paragraph{Case 2: Diagnostic uncertainty under severe signal distortion.}

The second anonymized DiagnosisToken described the gearbox condition of turbine WT-B over the same analysis period. The data observability was 100\%; however, severe signal distortion occurred on all 31 monitoring days. The distortion frequency was therefore 100\%, with a maximum consecutive duration of 31 days. Although no fault indication was generated, the fault-diagnosis conclusion was classified as unreliable because of the persistent signal distortion.

The textual DiagnosisToken provided to the LLM was formulated as follows:

\begin{quote}
Turbine WT-B, gearbox, left radial measurement position, was monitored over a one-month period with complete data observability. Severe signal distortion occurred on all 31 monitoring days, resulting in a distortion frequency of 100\% and a maximum consecutive duration of 31 days. Although no fault indication was produced, the fault-diagnosis result was considered unreliable because of the persistent signal distortion. Inspection of the vibration monitoring system was recommended before further assessment of gearbox health.
\end{quote}

For this case, the LLM was asked to determine whether the absence of a fault indication could be interpreted as evidence of normal gearbox health, explain the influence of signal distortion on diagnostic confidence, and identify the appropriate maintenance priority.


The generated responses were retained without manual modification. These examples are intended to demonstrate the feasibility of using structured diagnostic text as an intermediate interface between an industrial analytical subsystem and an LLM, rather than to provide a comprehensive evaluation of LLM decision accuracy.
\section{Discussion and Future Work}
\label{sec:future_work}

The present study provides an initial implementation of Industrial Tokenization through an end-to-end DiagnosisToken pathway. Record-level vibration-diagnosis results are aggregated into structured textual representations and subsequently interpreted by an LLM. This implementation demonstrates how the output of an autonomous industrial analytical subsystem can be connected to an LLM through a language-based intermediate interface.

However, the current implementation is limited to a single diagnostic subsystem. Therefore, it does not yet constitute a complete multi-source federation. In addition, the DiagnosisTokens are generated using predefined aggregation and interpretation rules, and the LLM-based interpretation is illustrated using representative cases rather than evaluated through a comprehensive quantitative experiment.

Future work will introduce a SCADA-based condition-monitoring token and integrate it with the current DiagnosisToken pathway. This extension will enable the central LLM to compare and fuse evidence derived from vibration diagnosis and SCADA condition monitoring. Additional industrial analytical subsystems and more automated token-generation methods may subsequently be incorporated into the proposed federated architecture.

\section{Conclusion}
\label{sec:conclusion}

This paper introduces Industrial Tokenization as a language-based interface for connecting heterogeneous industrial analytical subsystems with LLM-based health intelligence. Instead of directly unifying raw industrial data or internal model representations, Industrial Tokenization transforms source-specific analytical outputs into relatively standardized, sufficiently descriptive, and LLM-readable Industrial Tokens.

Based on this concept, a Federated Industrial Architecture is proposed, in which autonomous analytical subsystems retain their own data structures, models, and domain-specific processing methods, while their outputs are connected through a unified token interface to a central LLM-based fusion layer. The architecture is intended to support the integration of heterogeneous industrial evidence without requiring the underlying analytical subsystems to be homogeneous.

An initial end-to-end DiagnosisToken pathway is implemented using automated vibration-based diagnosis, rule-based event aggregation, structured textual token generation, and LLM-based interpretation. This implementation demonstrates how
record-level diagnostic results can be transformed into a compact intermediate representation that preserves equipment identity, signal quality, fault evidence, temporal persistence, confidence information, and maintenance-oriented
recommendations.

The present study is limited to a single diagnostic subsystem and illustrative LLM interpretation cases. Future work will extend the framework by introducing additional Industrial Tokens, including SCADA-based condition-monitoring tokens, to support cross-source evidence fusion and more comprehensive equipment health assessment.

\bibliographystyle{unsrt}  

\bibliography{references}

@article{FENG20124919,
title = {Vibration signal models for fault diagnosis of planetary gearboxes},
journal = {Journal of Sound and Vibration},
volume = {331},
number = {22},
pages = {4919-4939},
year = {2012},
issn = {0022-460X},
doi = {https://doi.org/10.1016/j.jsv.2012.05.039},
url = {https://www.sciencedirect.com/science/article/pii/S0022460X12004415},
author = {Zhipeng Feng and Ming J. Zuo}
}

@article{WANG2006176,
title = {Vibration-based fault diagnosis of pump using fuzzy technique},
journal = {Measurement},
volume = {39},
number = {2},
pages = {176-185},
year = {2006},
issn = {0263-2241},
doi = {https://doi.org/10.1016/j.measurement.2005.07.015},
url = {https://www.sciencedirect.com/science/article/pii/S0263224105001302},
author = {Jiangping Wang and Hongtao Hu}
}

@article{JARDINE20061483,
title = {A review on machinery diagnostics and prognostics implementing condition-based maintenance},
journal = {Mechanical Systems and Signal Processing},
volume = {20},
number = {7},
pages = {1483-1510},
year = {2006},
issn = {0888-3270},
doi = {https://doi.org/10.1016/j.ymssp.2005.09.012},
url = {https://www.sciencedirect.com/science/article/pii/S0888327005001512},
author = {Andrew K.S. Jardine and Daming Lin and Dragan Banjevic}
}

@article{ZHANG2024108236,
title = {Multi-modal data cross-domain fusion network for gearbox fault diagnosis under variable operating conditions},
journal = {Engineering Applications of Artificial Intelligence},
volume = {133},
pages = {108236},
year = {2024},
issn = {0952-1976},
doi = {https://doi.org/10.1016/j.engappai.2024.108236},
url = {https://www.sciencedirect.com/science/article/pii/S0952197624003944},
author = {Yongchao Zhang and Jinliang Ding and Yongbo Li and Zhaohui Ren and Ke Feng}
}

@article{KONG2020760,
title = {Condition monitoring of wind turbines based on spatio-temporal fusion of SCADA data by convolutional neural networks and gated recurrent units},
journal = {Renewable Energy},
volume = {146},
pages = {760-768},
year = {2020},
issn = {0960-1481},
doi = {https://doi.org/10.1016/j.renene.2019.07.033},
url = {https://www.sciencedirect.com/science/article/pii/S0960148119310535},
author = {Ziqian Kong and Baoping Tang and Lei Deng and Wenyi Liu and Yan Han}
}

@ARTICLE{1546063,
  author={Nandi, S. and Toliyat, H.A. and Li, X.},
  journal={IEEE Transactions on Energy Conversion}, 
  title={Condition Monitoring and Fault Diagnosis of Electrical Motors—A Review}, 
  year={2005},
  volume={20},
  number={4},
  pages={719-729},
  doi={10.1109/TEC.2005.847955}}

@ARTICLE{10960316,
  author={Han, Honggui and Meng, Yuan and Wu, Xiaolong and Li, Xin and Qiao, Junfei},
  journal={IEEE Transactions on Instrumentation and Measurement}, 
  title={A Transfer Learning-Based Multimodal Feature Fusion Model for Bearing Fault Diagnosis}, 
  year={2025},
  volume={74},
  number={},
  pages={1-13},
  doi={10.1109/TIM.2025.3558745}}

@ARTICLE{11580333,
  author={Deshpande, Pratik Anand and Preetha Roselyn, J. and Sundaravadivel, Prabha},
  journal={IEEE Access}, 
  title={Development of Feature Tokenizer Deep Learning Model for Fault Diagnosis in Marine Propulsion System}, 
  year={2026},
  volume={14},
  number={},
  pages={99482-99500},
  doi={10.1109/ACCESS.2026.3707741}}

@article{LI2025103277,
title = {HSE: A plug-and-play module for unified fault diagnosis foundation models},
journal = {Information Fusion},
volume = {123},
pages = {103277},
year = {2025},
issn = {1566-2535},
doi = {https://doi.org/10.1016/j.inffus.2025.103277},
url = {https://www.sciencedirect.com/science/article/pii/S1566253525003501},
author = {Qi Li and Bojian Chen and Qitong Chen and Xuan Li and Zhaoye Qin and Fulei Chu}
}

@INPROCEEDINGS{10336112,
  author={Wang, Huan and Li, Yan-Fu},
  booktitle={2023 5th International Conference on System Reliability and Safety Engineering (SRSE)}, 
  title={Large Language Model Empowered by Domain-Specific Knowledge Base for Industrial Equipment Operation and Maintenance}, 
  year={2023},
  volume={},
  number={},
  pages={474-479},
  doi={10.1109/SRSE59585.2023.10336112}}

@ARTICLE{10557154,
  author={Wang, Huan and Li, Chenxi and Li, Yan-Fu and Tsung, Fugee},
  journal={IEEE Transactions on Industrial Cyber-Physical Systems}, 
  title={An Intelligent Industrial Visual Monitoring and Maintenance Framework Empowered by Large-Scale Visual and Language Models}, 
  year={2024},
  volume={2},
  number={},
  pages={166-175},
  doi={10.1109/TICPS.2024.3414292}}

@article{JIN2022379,
title = {A Time Series Transformer based method for the rotating machinery fault diagnosis},
journal = {Neurocomputing},
volume = {494},
pages = {379-395},
year = {2022},
issn = {0925-2312},
doi = {https://doi.org/10.1016/j.neucom.2022.04.111},
url = {https://www.sciencedirect.com/science/article/pii/S0925231222005112},
author = {Yuhong Jin and Lei Hou and Yushu Chen}
}

@article{ZHANG2025125861,
title = {LLM-TSFD: An industrial time series human-in-the-loop fault diagnosis method based on a large language model},
journal = {Expert Systems with Applications},
volume = {264},
pages = {125861},
year = {2025},
issn = {0957-4174},
doi = {https://doi.org/10.1016/j.eswa.2024.125861},
url = {https://www.sciencedirect.com/science/article/pii/S0957417424027283},
author = {Qi Zhang and Chao Xu and Jie Li and Yicheng Sun and Jinsong Bao and Dan Zhang}
}

@misc{HE202412481,
  title={MaintAGT: Sim2Real-Guided Multimodal Large Model for Intelligent Maintenance with Chain-of-Thought Reasoning},
  author={He, Hongliang and Huang, Jinfeng and Li, Qi and Wang, Xu and Zhang, Feibin and Yang, Kangding and Meng, Li and Chu, Fulei},
  year={2024},
  eprint={2412.00481},
  archivePrefix={arXiv},
  primaryClass={cs.AI}
}

@ARTICLE{10856008,
  author={Abdullahi, Shamsu and Usman Danyaro, Kamaluddeen and Zakari, Abubakar and Abdul Aziz, Izzatdin and Amila Wan Abdullah Zawawi, Noor and Adamu, Shamsuddeen},
  journal={IEEE Access}, 
  title={Time-Series Large Language Models: A Systematic Review of State-of-the-Art}, 
  year={2025},
  volume={13},
  number={},
  pages={30235-30261},
  doi={10.1109/ACCESS.2025.3535782}}

@ARTICLE{11509648,
  author={Wang, Yuxuan and Wu, Haixu and Dong, Jiaxiang and Liu, Yong and Wang, Chen and Long, Mingsheng and Wang, Jianmin},
  journal={IEEE Transactions on Pattern Analysis and Machine Intelligence}, 
  title={Deep Time Series Models: A Comprehensive Survey and Benchmark}, 
  year={2026},
  volume={},
  number={},
  pages={1-20},
  doi={10.1109/TPAMI.2026.3690845}}

\end{document}